\documentclass[12pt]{article}   	% use "amsart" instead of "article" for AMSLaTeX format
\usepackage{geometry}                		% See geometry.pdf to learn the layout options. There are lots.
\usepackage[noframe]{showframe}
\usepackage{graphicx}				% Use pdf, png, jpg, or eps§ with pdflatex; use eps in DVI mode
\usepackage{amssymb}
\usepackage{amsmath}
\usepackage{amsthm}
\usepackage[]{algorithm2e}
\usepackage{array}
\usepackage{mathtools}
\allowdisplaybreaks

\usepackage{afterpage}
\usepackage{threeparttable}
\usepackage{tikz}
\usepackage{emptypage}

\usepackage{longtable}
\usepackage{booktabs}
\newcommand{\ra}[1]{\renewcommand{\arraystretch}{#1}}
\usepackage{arydshln}
\usepackage{makecell}
\usepackage{pbox}
\usepackage{epstopdf}

\usepackage{subcaption}
\usepackage{parskip}
\usepackage{array}
\newcolumntype{P}[1]{>{\centering\arraybackslash}p{#1}}
\newcolumntype{M}[1]{>{\centering\arraybackslash}m{#1}}
\usepackage{multirow}
\usepackage{bm}
\newcommand{\bal}{\bm{\alpha}}
\newcommand{\norm}[1]{\left\lVert#1\right\rVert}

\usepackage[utf8]{inputenc}
\usepackage[english]{babel}

\usepackage{natbib}
\setcitestyle{authoryear,open={(},close={)}}

\usepackage{csquotes}
\renewcommand{\mkbegdispquote}[2]{\itshape}

\usepackage{authblk}

\usepackage[toc, page]{appendix}

\usepackage{bbm}
\usepackage{fancyhdr}
\title{Exclusive Topic Modeling}
\author[1]{Hao LEI }
\author[2,3]{Ying CHEN }
\affil[1]{Department of Statistics and Applied Probability, National University of Singapore}
\affil[2]{Department of Mathematics, National University of Singapore}
\affil[3]{Risk Management Institute, National University of Singapore}

\date{}
\begin{document}
\maketitle
\begin{abstract}
	We propose an Exclusive Topic Modeling (ETM) for unsupervised text classification, which is able to 1) identify the field-specific keywords though less frequently appeared and 2) deliver well-structured topics with exclusive words. In particular, a weighted Lasso penalty is imposed to reduce the dominance of the frequently appearing yet less relevant words automatically, and a pairwise Kullback-Leibler divergence penalty is used to implement topics separation. Simulation studies demonstrate that the ETM detects the field-specific keywords, while LDA fails. When applying to the benchmark NIPS dataset, the topic coherence score on average improves by $22\%$ and $10\%$ for the model with weighted Lasso penalty and pairwise Kullback-Leibler divergence penalty, respectively.
\end{abstract}

\section{Introduction}

Topic modeling has been widely used in many different fields, including scientific topic extraction \citep{blei2007correlated}, cryptocurrency \citep{CRY}, operation risk extraction \citep{huang2017analyst}, communication research \citep{maier2018applying},  marketing \citep{reisenbichler2019topic}, investor attention modeling \citep{lei2020investor}, computer vision \citep{FEI}, bio-informatics \citep{liu2016overview, gonzalez2019cistopic}. Two well-known challenges in topic modeling are: 1)the predominance  of the frequently appearing words in the estimated topics; 2) topics are overlapped with common words, making the structure and interpretation difficult. We propose an Exclusive Topic Model (ETM) to tackle these two issues. ETM can identify field-specific keywords and deliver well-structured topics with exclusive words. More specifically, a weighted Lasso penalty is imposed to reduce the predominance of the frequently appearing yet less relevant words automatically and a pairwise Kullback-Leibler divergence penalty is used to implement topics separation. 

Topic modeling makes use of the word co-occurrence information to estimate topics. Due to the human language habit and structure, certain words appear more frequently than others, e.g. the Zipf's law. Consequently, the frequently appearing words co-occur with more words and thus are predominant in the estimated topics. The phenomenon makes topic interpretation difficult, as general and frequently appearing words take the place of the true exclusive topic words. The semantic coherence of the estimated topics also deteriorates. Researchers have developed a couple of models to tackle the challenge. \cite{wallach2009rethinking} propose to use asymmetric Dirichlet prior to alleviate the predominance of frequently appearing words. To take account of the uncertainty in the parameters of asymmetric Dirichlet prior, they add a hyper prior to the prior parameters, which are then integrated out during the  estimation. \cite{griffiths2005integrating} propose an LDA-HMM (Hidden Markov Model) model, which separates the short-term dependent syntactic and long-term dependent semantic words into different topics. The separation mitigates the predominance of syntactic words, but not the semantic words. Several methods to measure the topic coherence and word intrusion. 

Topic models make assumptions on the topic and word distributions. In reality, these assumptions are not fully satisfied. The violation sometimes makes the estimated topics having similar semantic meanings and share several common words. This issue is related to the topic number selection, as choosing too many topics will result in many similar small topics \citep{greene2014many}. \cite{LDA} use cross-validation to select the number of topics that produces the smallest perplexity. \cite{griffiths2004hierarchical} add the Chinese Restaurant Process as a prior for the number of topics to automatically find the number of topics. In practice, it usually results in too many topics. \cite{greene2014many} propose a term-centric stability analysis strategy. 
Researchers also try to improve the topic-word distributions by employing other information or adding new latent variables. \cite{rabinovich2014inverse} separate the topic-word distribution into a base distribution and a document-specific parameter that serves to distort the base distribution for a better fit. \cite{das2015gaussian, shi2017jointly, xu2018distilled} use word embedding information from the neural network to improve the topics. 

Lasso performs variable selection and regularization in regression analysis \citep{tibshirani1996regression}. Its weighted version provides more flexibility as different penalties can be applied to different parameters. The weighted lasso has been applied in various fields. \cite{shimamura2007weighted} propose weighted lasso estimation for the graphical Gaussian model of large gene networks from DNA microarray data. The weighted lasso is flexible to add different penalties in the neighborhood selection of the graphical Gaussian model. \cite{angelosante2009rls} develop a weighted version of the recursive Lasso with weights obtained from the recursive least square algorithm. They show that the weighted Lasso algorithm estimate sparse signals consistently. \cite{park2013lag} propose a lag weighted Lasso for the time series model, where the weights reflect both the penalty size and the lag effect. Simulation and real data show that the proposed method is superior to both lasso and adaptive lasso in forecasting accuracy. \cite{zhao2015wavelet} show that weighted lasso leads to improved estimation and prediction than lasso in wavelet functional linear regression. Weighted lasso has also been applied with geographical data. For example, \cite{wang2020assessing} use geographically weighted lass to assess geochemical anomalies; \cite{he2020adapted} use the geographically weighted lasso to predict the subway ridership. 

Kullback-Leibler divergence is often used as a measure of closeness between two probability distributions. It often appears in machine learning, especially the variational inference literature as maximizing the Evidence Lower BOund(ELBO) is equivalent to minimizing the KL divergence between the mean-field variational distribution and the true posterior \citep{blei2017variational}. Besides, it is also used in many other fields. \cite{smith2006markov} develop a criterion, which is an estimate of the KL divergence of the true and candidate models, for the number of states and variables selection in the Markov switching models.  
\cite{gupta2009classification} combine Kullback-Leibler divergence with KNN, in which KL divergence is used as a distance measure, and SVM, in which KL divergence is used as kernels. They show that these combinations produce favorable results comparing to the Euclidean KNN and SVM with linear and radial basis functions in classifying the electroencephalography signals. \cite{hsu2015neural} propose a neural network framework for classification which is trained using weak labels, i.e. the pairwise relationships between data instances. In the framework, they replace the usual cross-entropy cost function with the pairwise KL divergence, which takes into the neural network output and the pairwise relationships between the training data instances. \cite{lin2018model} minimize the KL divergence to learn the coupling parameters in the Ising or Heisenberg spin configurations with the Boltzmann type distribution.          
\cite{lu2019fault} maximize the pairwise ratio Kullback-Leibler divergence in their industrial process fault diagnosis.

We propose an Exclusive Topic Model (ETM) which tackles the predominance of frequently appearing words in the estimated topics and the 'close' topic challenges. More specifically, a weighted Lasso penalty is used to penalize the frequently appearing words during the topic estimation. Different weights reflect the inherited appearing frequencies of different words. As a result, frequently appearing words are penalized more during the estimation and the predominance is mitigated in the estimators. A pairwise KL divergence penalty is added to separate the topics. In this case, a linear combination of ELBO and the penalty is jointly maximized. The estimator achieves the balance between these two terms. We develop the variational EM algorithms for the proposed model. Simulation studies and the public available NIPS dataset are used to demonstrate the effectiveness of the prosed method. The simulation studies show that 1) ETM can effectively mitigate the predominance of the frequently appearing words in the estimated topics; 2) ETM can be used to incorporate the prior information to discover the important but infrequently appearing words in the corpus; 3) ETM is able to separate the close topics, which share common words. Applying the ETM on the public available NIPS dataset, the topic coherence of ETM improves by $22\%$ and $10\%$ for the weighted lasso penalty and the pairwise KL divergence penalty, respectively.

\iffalse
We find two papers sharing similar regularization idea as ours. \cite{newman2011improving} use quadratic regularizer and convolved Dirichlet regularizer which utilizes the short-range word dependency information to improve topic coherence measured by PMI scores. \cite{vorontsov2015additive} use entropy regularizer to eliminate insignificant and linearly dependent topics.  Comparing to those two papers, our method uses different penalty terms and achieves different goals. Our weighted LASSO penalty is used to incorporate prior information about the topics, and the KL divergence penalty is to increase the 'distance' among the topics.

Our main contribution is to proivde a penalty method to tackle two well-known issues in the topic modeling community: 1) frequently appearing words dominant the discovered topics; 2) some discovered topics are 'close' to each other. We emphasize that the penalty method is more general than only tackling these two issues. The weights in the penalties can be set to incorporate prior information, which could potentially improve the topic interpretation and the accuracy of the estimators.
\fi

The rest of this paper is organized in the following way. In section \ref{sec:plda}, we provide details of the proposed method and the algorithms to estimate the topics.  In section \ref{sec:simulation}, we conduct three simulation studies to demonstrate the effectiveness of the proposed method in tackling the two common issues in topic modeling.  In section \ref{sec:nips}, we apply the proposed method to the public NIPS dataset. The results show that the proposed method improves topic interpretability and coherence scores. We conclude the paper in section \ref{sec:conclusion}.

\section{Method} \label{sec:plda}
In this section, we provide the details of the proposed methods. In section \ref{sec:etm}, ETM is described on a high level. Since there are two penalties, we study their corresponding effect on the topic estimation separately. In section \ref{sec:wl}, we show the details of adding the weighted LASSO penalty and the updating equations. We give a modified variational m-step algorithm for the weighted LASSO penalty. In section \ref{sec:kld}, we present the details of adding the pairwise Kullback-Leibler divergence penalty. Different from the weighted LASSO penalty, the objective function is no longer convex. We use an algorithm that combines gradient descent and Hessian descent to find updating equations. In section \ref{sec:dpw}, we discuss how to combine them and implement them in practice.

\subsection{ETM} \label{sec:etm}
The model set up is the same as LDA \citep{LDA}. Namely, given a corpus $C$, we assume it contains $K$ topics. Every topic $\eta_k$ is multinomial distribution on the vocabulary. Every document $d$ contains one or more topics. The topic proportion in each document is governed by the local latent parameter document-topic $\theta$, which has a Dirichlet prior with hyperparameter $\zeta$.  Every word in document $d$ is generated from the contained topics as follows:

\begin{itemize}
	\item for every document $d \in C$, its topic proportion parameter $\theta$ is generated from a Dirichlet distribution, i.e. $\theta \sim Dir(\zeta)$.
	\item for every word in the document $d$,
	\begin{itemize}
		\item a topic $Z$ is first generated from the multinomial distribution with parameter $\theta$, i.e. $Z \sim Multinomial(\theta)$
		\item a word $w$ is then generated from the multinomial distribution with parameter $\eta_Z$, i.e. $w \sim Multinomial(\eta_Z)$
	\end{itemize}
\end{itemize}

 \begin{figure}[htbp]
	\begin{center}
		\includegraphics[width=\textwidth, height=0.4\textheight]{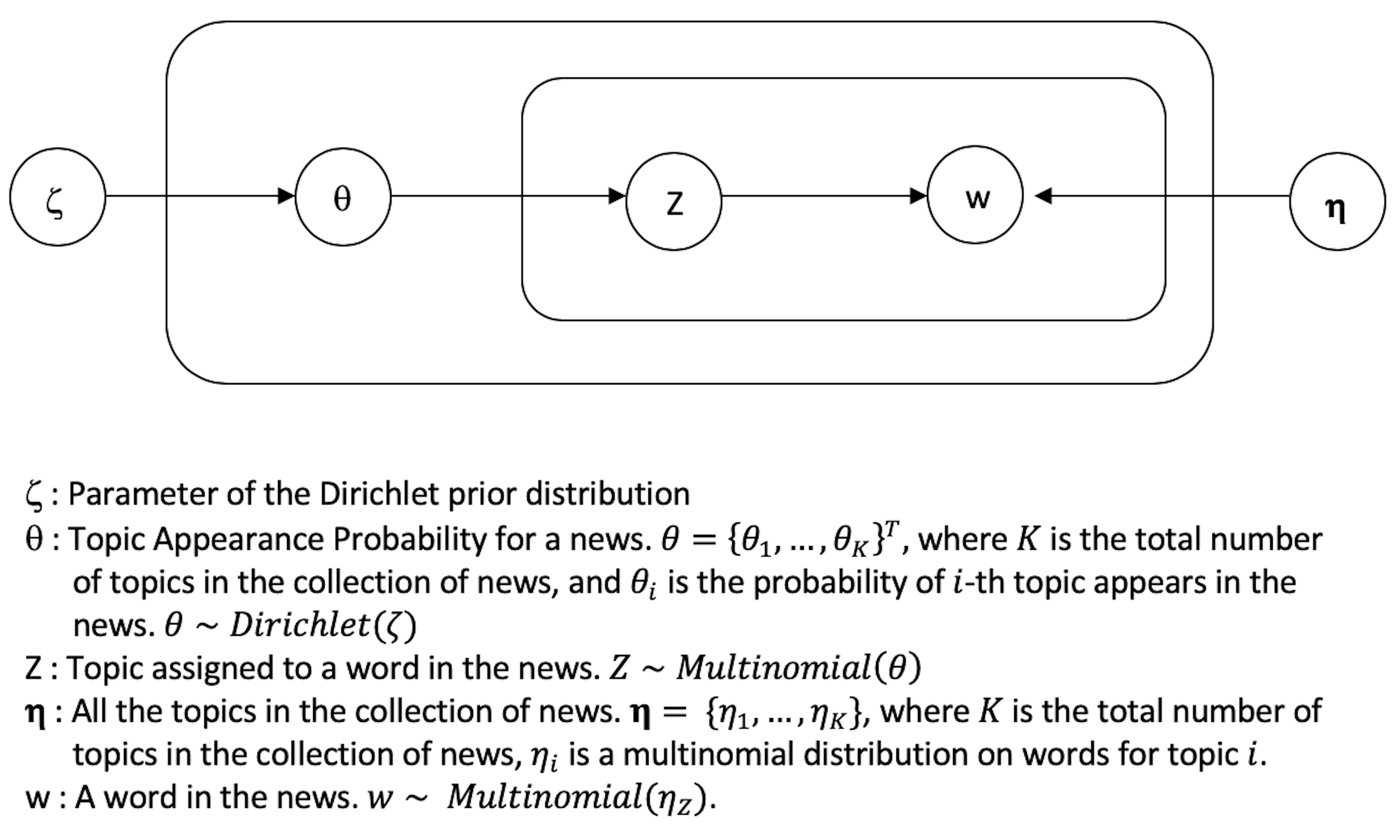}
		\caption{ The graphical representation. The outer box represents the document level. The inner rectangle represents the word level. }
		\label{fig:lda}
	\end{center}
\end{figure}

The graphical representation is shown in Figure \ref{fig:lda}. The outer rectangle represents the document-level and the inner rectangle represents the word-level. $\zeta$ and $\eta$ are global parameters, i.e. shared by all the documents. $\theta$ and $Z$ are local latent variables. A complete Bayesian approach further assumes that topics $\eta_1, \dots, \eta_K$  are generated from a Dirichlet prior with hyperparameter $\beta$. Here we use this formulation as in \cite{LDA} for the ease of adding a penalty. 

The latent parameters in ETM are estimated by maximizing the following penalized posterior.
\begin{equation}
	\max \hspace{2mm}  p(\theta, Z,  \zeta, \eta | W ) - \sum_{i=1}^{K} \sum_{j=1}^{V} \mu_i m_{ij} |\eta_{ij}| + \sum_{i=1, l \ne i}^{K}  \nu_{il} D_{KL}(\eta_i || \eta_l)
\end{equation}
where $p(\theta, Z,  \zeta, \eta | W )$ is the posterior. The second term is the weighted lasso penalty, in which $\mu_i$ is the penalty weight for topic $i$, $m_{ij}$ represents the weight for word $j$ in topic $i$ and is known. The third term is the pairwise KL divergence penalty, in which $D_{KL}(\eta_i || \eta_l)$ represents the KL divergence between topic $i$ and $l$ and $\nu_{il}$ is the corresponding penalty weight. 

\iffalse
In the model, the only observables are the words $w$. The latent variables are estimated by maximizing the posterior

	\begin{equation*}
		p(\bm{Z},\bm{\theta}|\bm{W},\bal,\bm{\beta}) = \frac{p(\bm{Z},\bm{\theta},\bm{W}|\bal,\bm{\beta}) }{p(\bm{W}|\bal,\bm{\beta}) }
	\end{equation*}
\fi
Unfortunately, the posterior is intractable to compute. Instead, a `variational EM algorithm' is used to maximize the Evidence Lower Bound (ELBO) \citep{LDA, blei2017variational}  
 $L( \zeta, \eta, \gamma, \phi)$,
 \begin{align*}
 	 L( \zeta, \eta, \gamma, \phi) &=  E_q[ \ln p(\theta, Z, W | \zeta, \eta)]-E_q[\ln q(\theta,Z| \gamma, \phi) ] \\ 
 	 & \le  \ln p( W| \zeta, \eta)
 \end{align*}
where $p(.)$ is the density function derived from LDA and $q(\theta, Z | \gamma, \phi)$ is the mean-field variational distribution 
\begin{equation*}
	q(\theta, Z | \gamma, \phi) = q(\theta | \gamma)\prod_{n=1}^{N} q(Z_n|\phi_n)
\end{equation*}
where $N$ is the number of words in a document, $q(\theta | \gamma) \sim Dirichlet(\gamma)$, and $q(Z_n|\phi_n) \sim Multinomial(\phi_n)$. $E_q$ represents the expectation under the variational distribution. The inequality is a result of applying Jensen inequality. The data $W$ provides more evidence to our prior belief. Hence the name ELBO. 
 
 \iffalse
    \begin{align*}
 	\ln p( \bm{W}| \bal, \bm{\beta}) %&= \ln \int \sum_{\bm{Z}} p(\bm{\theta}, \bm{Z}, \bm{W} | \bal, \bm{\beta}) d\bm{\theta} \\ 
 	%&= \ln \int \sum_{\bm{Z}} \frac{p(\bm{\theta}, \bm{Z}, \bm{W} | \bal, %\bm{\beta})}{q(\bm{\theta},\bm{Z}| \gamma, \phi)} q(\bm{\theta},\bm{Z}| \gamma, %\phi)d\bm{\theta} \\ 
 	%&= \ln E_q[ \frac{p(\bm{\theta}, \bm{Z}, \bm{W} | \bal, \bm{\beta})}{q(\bm{\theta},\bm{Z}| %\gamma, \phi)} ] \\ 
 	& \ge  E_q[ \ln p(\bm{\theta}, \bm{Z}, \bm{W} | \bal, \bm{\beta})]-E_q[\ln q(\bm{\theta},\bm{Z}| \gamma, \phi) ] \\ 
 	& = L( \bal, \bm{\beta},\gamma, \phi)
 \end{align*}
  which is the difference between the log-posterior and the Kullback-Leibler divergence between the true distribution and the mean-field variational family . In the E-step, we maximize the local latent variable $\theta$ and $Z$ for every document and word, with updating equation
 \fi
 
 Therefore, in the actual optimization, we maximize the following penalized ELBO.
 \begin{equation} \label{equ:pen_elbo}
 	\max \hspace{2mm}   L( \zeta, \eta, \gamma, \phi)  - \sum_{i=1}^{K} \sum_{j=1}^{V} \mu_i m_{ij} |\eta_{ij}| + \sum_{i=1, l \ne i}^{K}  \nu_{il} D_{KL}(\eta_i || \eta_l)
 \end{equation}
Equation \ref{equ:pen_elbo} is maximized in an `EM'-like procedure. In the E-step, the ELBO is maximized w.r.t. the local variational parameter $\phi, \gamma$ for every document, conditional on the global latent parameter $\eta, \zeta$. Since the penalties don't contain any local variational parameters, the updating equations will be the same as that of LDA.
\begin{equation} \label{equ:e-equ}
	\begin{aligned}
		& \phi_{ni} \propto \eta_{iw_n}\exp{E_q[\log (\theta_i) | \gamma]} \\
		& \gamma_i = \zeta_i + \sum_{n=1}^N \phi_{ni}
	\end{aligned}
\end{equation}
where
\begin{equation*}
	\exp{E_q[\log (\theta_i) | \gamma]} = \Psi(\gamma_i) - \Psi(\sum_{l=1}^K \gamma_l)
\end{equation*}
and $\Psi$ is the digamma function, i.e. the logarithmic derivative of the gamma function.

Then conditional on all the local latent variables $\phi, \gamma$, the penalized ELBO is maximized w.r.t. the latent global parameter $\eta, \zeta$. The global parameter $\zeta$ can be estimated using Newton's method. In practice, $\zeta$ is often assumed to be a symmetric Dirichlet parameter.

\subsection{Only Weighted Lasso Penalty: $\nu = 0$} \label{sec:wl}
In this section, we consider the case where we only have the weighted lasso penalty, i.e. $\nu = 0$. 
\begin{equation} \label{equ:md}
	\max \hspace{2mm} L( \zeta, \eta, \gamma, \phi) - \sum_{i=1}^{K} \sum_{j=1}^{V} \mu_i m_{ij} |\eta_{ij}|
\end{equation}
subject to
\begin{equation*}
	\begin{aligned}
		& \sum_{j=1}^{V} \eta_{ij} = 1, \forall i \in \{1, \dots, K\} \\
		& \eta_{ij} \ge 0, \forall i, j \\
		& \sum_{i=1}^{K} \theta_{di} = 1, \forall d
	\end{aligned}
\end{equation*}
where $K$ and $V$ represent the number of topics and vocabulary size respectively and $\mu_i \ge 0$ is the penalty weight for topic $i$ and is selected using cross-validation, $\eta_{ij}$ represents the probability of the $j$th word in the topic $i$, $m_{ij}$ is the weight for $\eta_{ij}$ and is known in advance, reflecting the prior information about the topic-word distribution. One possible candidate for the weight is the document frequency, i.e. $m_{ij} = df_{j}, \forall i, j$, where $df_j$ is the number of documents containing the word $j$. The larger the document frequency for the word $j$, the larger the penalty. Consider an extreme case that word $j$ appears in every document of the corpus. Due to it co-occurs with every other word, LDA would assign a large probability to it in every topic. As a result, word $j$ contains little information to distinguish one topic from another. It's barely useful in the dimension reduction process, i.e. from word space to topic space. With the document frequency penalty, it will be penalized the most and result in a low probability in the topic distributions.

We emphasize that the penalized model is not constrained to only solving the frequently words dominance issue. Any weight reflecting the prior information about the topic distribution can be used to achieve the practitioners' goal. We give an illustration here and in the simulation study \ref{sec:sim2}. Often practitioners found the field-important words are not assigned large probabilities in the estimated topics. (One possible reason is that they appear infrequently in the underlying corpus.) But these keywords contain important information about the field and are crucial to distinguish one topic from another. And practitioners might prefer they appear in the top-$T$ words for easy topic interpretation (In practice, $T$ is usually set as $10$ or $20$). In this situation, practitioners can utilize our proposed model by assigning negative weights to these keywords and zero weights to all the other words. Simulation case \ref{sec:sim2} is devoted to the situation.  

\begin{figure}[htbp]
	\begin{center}
		\includegraphics[width=\textwidth, height=0.45\textheight]{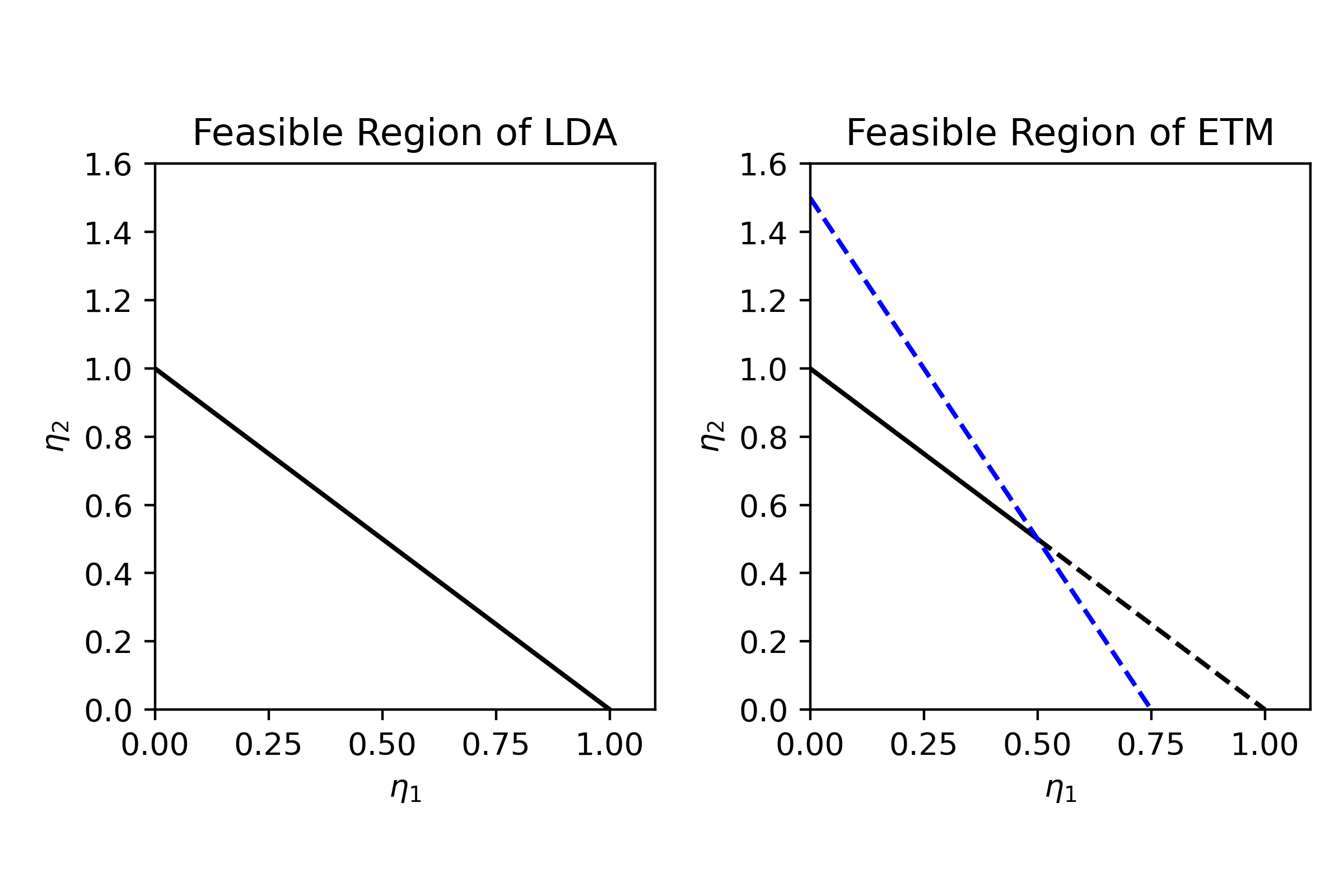}
		\caption{ The feasible region of topics of LDA and ETM. The vocabulary consists of two words $w_1$ and $w_2$. Their corresponding document frequencies are 2 and 1, respectively. The feasible region of LDA topics is plotted in the left subplot as the solid black line. That of df-weighted LASSO penalized LDA is plotted in the right subplot. The blue dashed line is the penalty induced constraint line $2\eta_1 + \eta_2 = 1.5$. The extra constraint reduces the feasible region of the topics to the top left black solid line. As a result, the feasible probability range $\eta_1$ is reduced to $(0, 0.5)$. A smaller weight will be assigned to the relatively more frequently appearing word $w_1$ in the estimated topic.}
		\label{fig:feasible_region}
	\end{center}
\end{figure}

To further understand the penalization, we rewrite the optimization problem \ref{equ:md} in the following equivalent form.
\begin{equation*}
\max \hspace{2mm} L( \zeta, \eta, \gamma, \phi)
\end{equation*}
subject to
\begin{equation*}
\begin{aligned}
& \sum_{j=1}^{V} m_{ij} |\eta_{ij}| \le \nu_i, \forall i \in \{1, \dots, K\} \\
& \sum_{j=1}^{V} \eta_{ij} = 1, \forall i \in \{1, \dots, K\} \\
& \eta_{ij} \ge 0, \forall i, j \\
& \sum_{i=1}^{K} \theta_{di} = 1, \forall d
\end{aligned}
\end{equation*}
where $\nu_i > 0$ is a hyperparameter and there is a one-to-one correspondence between $\mu_i$ and $\nu_i$. The penalty alters the feasible region. Figure \ref{fig:feasible_region} plots the feasible regions of the topics of LDA and the df-weighted LASSO penalized LDA for a simple case of having two words $w_1$ and $w_2$. The document frequencies are 2 and 1 for the word $w_1$ and $w_2$, respectively. The black solid line in the left subplot represents the feasible region of LDA. The feasible region of the ETM is plotted in the right subplot. The blue dashed line is the penalty induced constraint line $2\eta_1 + \eta_2 = 1.5$. Due to the extra constraint, the feasible region is reduced to the upper-left black solid line. As a result, the feasible probability range $\eta_1$ is reduced to $(0, 0.5)$. A smaller weight will be assigned to the relatively more frequently appearing word $w_1$ in the estimated topic.

%The model assumes that the news corpus contains $K$ topics, $\zeta_1, \dots, \zeta_K$. Each topic has a multinomial distribution over the vocabulary. Every news contains one or several topics of the total $K$ topics. The proportion of each topic in news is represented by $\theta$. $\theta$ is generated from a Dirichlet distribution with parameter $\zeta$, i.e. $\theta \sim \text{Dir}(\zeta)$. The $\zeta$ is a hyper-parameter, representing the prior information of how likely each topic occurs. We use a non-informative prior and let $\zeta = 1 / K$, meaning every topic is equally likely to occur in a news. For each word in the news, we first draw a topic $Z$ from a multinomial distribution with parameter $\theta$, i.e. $Z \sim \text{Multinomial}(\theta)$. Then we draw a word $w$ from the topic $Z$, i.e. $w \sim \text{Multinomial}(\eta_Z)$. Given the news collection, we estimate $\theta, Z, \eta$ by maximizing the posterior using variational inference \citep{LDA}.

As mentioned in section \ref{sec:etm}, the optimization is done using variational inference and the local latent parameters are updated the same as LDA, as given in equation \ref{equ:e-equ}. The global latent parameter $\eta$ is estimated by maximizing the following equation
\begin{equation*}
	\max \hspace{2mm} -f(\eta) = \sum_{d=1}^D\sum_{n=1}^{N_d}\sum_{i=1}^K \sum_{j=1}^V \phi_{dni}w_{dn}^j \log(\eta_{ij})  - \sum_{i=1}^{K} \sum_{j=1}^{V} \mu_i m_{ij} |\eta_{ij}|
\end{equation*}
subject to
\begin{equation*}
	\begin{aligned}
	& \sum_{j=1}^{V} \eta_{ij} = 1, \forall i \in \{1, \dots, K\} \\
	& \eta_{ij} \ge 0, \forall i, j
	\end{aligned}
\end{equation*}
where the first term is by taking out all the terms containing $\eta$ from ELBO.
The problem can be further reduced to $K$ sub-optimization problems below. Due to $\log (\eta_{ij})$ in the target function, the constraint $\eta_{ij} \ge 0, \forall i, j$ can be ignored. The absolute value $|\eta_{ij}|$ in the target function equals $\eta_{ij}$. We rewrite the maximization as an equivalent minimization problem.
\begin{equation} \label{equ:pw_min}
	\min \hspace{2mm} f_i(\eta_i) = - \sum_{d=1}^D\sum_{n=1}^{N_d} \sum_{j=1}^V \phi_{dni}w_{dn}^j \log(\eta_{ij})  +  \sum_{j=1}^{V} \mu_i m_{ij} \eta_{ij}
\end{equation}
subject to
\begin{equation*}
	\begin{aligned}
	& \sum_{j=1}^{V} \eta_{ij} = 1 \\
	\end{aligned}
\end{equation*}

We use the Newton method with equality constraints \citep{boyd2004convex} to solve equation \ref{equ:pw_min}. The updating direction $\Delta \eta_i$ with a feasible starting point $\eta_i^{0}$ can be calculated using the following equation
\begin{equation*}
	\begin{bmatrix}
	\nabla ^2f_i  & \mathbf{1} \\
	\mathbf{1}^T & 0

	\end{bmatrix}
	\begin{pmatrix}
	\Delta \eta_i \\
	\alpha_i
	\end{pmatrix}
	= \begin{pmatrix}
	-\nabla f_i \\
	0
	\end{pmatrix}
\end{equation*}
The updating direction is
\begin{equation}
	\Delta \eta_{ij} = \frac{- f^\prime_{ij} - \alpha_i }{f^{\prime \prime}_{ij}} = \eta_{ij} - \frac{(\alpha_i + \mu_im_{ij})\eta_{ij}^2}{\sum_{d=1}^{D}\sum_{n=1}^{N_d}\phi_{dni}w_{dn}^j}
\end{equation}
where $f^\prime_{ij}$ and $f^{\prime \prime}_{ij}$ are the first and second partial derivatives of $f_i$ with respect to $\eta_j$, and
\begin{equation*}
	\alpha_i = \frac{-\sum_{j=1}^V f^\prime_{ij}/f^{\prime \prime}_{ij}}{\sum_{j=1}^V 1 /f^{\prime \prime}_{ij}} = \frac{\sum_{j=1}^{V} \left( \eta_{ij} -\frac{ \mu_i m_{ij}\eta_{ij}^2}{\sum_{d=1}^{D}\sum_{n=1}^{N_d}\phi_{dni}w_{dn}^j}\right) }{\sum_{j=1}^{V}\frac{\eta_{ij}^2}{\sum_{d=1}^{D}\sum_{n=1}^{N_d}\phi_{dni}w_{dn}^j}}
\end{equation*}
The Newton decrement is
\begin{equation}
	\lambda(\eta_i) = (\Delta \eta_i^T \nabla^2f_i \Delta \eta_i)^{1/2}  = \left(\sum_{j=1}^{V}  \frac{ (\sum_{d=1}^{D}\sum_{n=1}^{N_d}\phi_{dni}w_{dn}^j - (\alpha_i + \mu_im_{ij})\eta_{ij})^2}{\sum_{d=1}^{D}\sum_{n=1}^{N_d}\phi_{dni}w_{dn}^j}  \right)^{1/2} 
\end{equation}
We use the backtracking line search \citep{boyd2004convex} to estimate the step size. The complete step to estimate topics of the weighted LASSO penalized LDA is given in algorithm \ref{pointwise_algo}. 

\begin{algorithm}[H] \label{pointwise_algo}
 \KwResult{Update the $i$th topic word distribution $\eta_i$ }
  Initialize $\eta_i$ with a feasible point\;
  Choose the stopping criteria $\epsilon$ and the line search parameter $\delta \in (0, 0.5)$, $\gamma \in (0, 1)$;

 \While{not reaching the maximum iteration}{
  Compute the feasible descent direction $\Delta \eta_i$ and Newton decrement $\lambda(\eta_i)$ \;
  \eIf{$\lambda(\eta_i)^2 / 2 \le \epsilon$}{
   Stop the algorithm ;
   }{\textbf{Find step size $t$ by backtracking line search:}\newline
	 Initialize the step size $t \coloneqq 1$ \;
   \While{$f_i(\eta_i + t\Delta \eta_i) > f_i(\eta_i) + \delta t \nabla f_i^T\Delta\eta$}{
	$ t \coloneqq \gamma t$;
	 }
  }
	$\eta_i \coloneqq \eta_i + t\Delta\eta_i$;
 }
\caption{Variational m-step}
\end{algorithm}

\subsection{Only Pairwise Kullback-Leibler Divergence Penalty: $\mu = 0$} \label{sec:kld}
Practitioners often find some estimated topics are 'close' to each other, in the sense, they have similar semantic meaning and share several common words in their top-$N$ words. It makes the topic interpretation and the analyzing steps following topic modeling, e.g. \cite{lei2020investor}, difficult. In this section, we consider the case where we only have a pairwise KL divergence penalty, i.e. $\mu = 0$. The optimization takes the following form. 
\begin{equation} \label{equ:pkl}
\max \hspace{2mm} L(\zeta, \eta, \gamma, \phi)  + \sum_{i=1, l \ne i}^{K}  \mu_{il} D_{KL}(\eta_i || \eta_l)
\end{equation}
subject to
\begin{equation*}
\begin{aligned}
& \sum_{j=1}^{V} \eta_{ij} = 1, \forall i \in \{1, \dots, K\} \\
& \eta_{ij} \ge 0, \forall i, j \\
& \sum_{i=1}^{K} \theta_{di} = 1, \forall d
\end{aligned}
\end{equation*}
where $	D_{KL}(\eta_i || \eta_l)$ is the KL divergence between topic $i$ and $l$
\begin{equation*}
	D_{KL}(\eta_i || \eta_l) = \sum_{j=1}^{V} \eta_{ij} \log (\frac{\eta_{ij}}{\eta_{lj}})
\end{equation*}

Since the penalty only involves topic distribution $\eta$ and thus only plays a role in the M-step. The E-step is the same as equation \ref{equ:e-equ}. For the M-step, we optimize
\begin{equation}
\min \hspace{2mm} g(\eta) = -\sum_{d=1}^D\sum_{n=1}^{N_d}\sum_{i=1}^K \sum_{j=1}^V \phi_{dni}w_{dn}^j \log(\eta_{ij})  - \sum_{i=1, l\ne i}^{K} \sum_{j=1}^{V} \mu_{il} \eta_{ij} \log (\frac{\eta_{ij}}{\eta_{lj}})
\end{equation}
subject to
\begin{equation*}
\begin{aligned}
& \sum_{j=1}^{V} \eta_{ij} = 1, \forall i \in \{1, \dots, K\} \\
& \eta_{ij} \ge 0, \forall i, j
\end{aligned}
\end{equation*}

There are two differences between the current optimization and the optimization in section \ref{sec:wl} : 1) the optimization can no longer be separated into $K$ sub-optimization problems as the different topics are now intertwined through the penalty; 2)the objective function is no long convex. For the first difference, we borrow the idea of coordinate descent and sequentially optimize one topic at a time while conditioning on all the other topics. The benefit of this approach instead of updating all topics simultaneously is that it simplifies the constrained Newton updating equation involving the Hessian matrix. Under the conditional approach, the Hessian matrix for a particular topic  $\nabla ^2 g_i$ is diagonal. For the second difference, due to the non-convexity, the Hessian matrix may not be positive semi-definite. As a result, Newton decrement could be a complex number. We use a combination of gradient descent and Hessian descent algorithm to tackle the second issue \citep{nesterov2006cubic, allen2018neon2}. The Hessian descent is invoked when the Hessian is not positive semi-definite. The Hessian descent moves to a smaller value along the Newton direction.   

We now show the updating equations for the gradient descent step. For topic $i$ while conditioning on all the other topics, we optimize the following equation 
\begin{equation} \label{equ:kl_per_topic}
\min \hspace{2mm} g_i(\eta_i | \eta_l, l \ne i) = - \sum_{d=1}^D\sum_{n=1}^{N_d} \sum_{j=1}^V \phi_{dni}w_{dn}^j \log(\eta_{ij})  -  \sum_{l \ne i} \sum_{j=1}^{V} \mu_{il} \eta_{ij} \log (\frac{\eta_{ij}}{\eta_{lj}}) - \sum_{l \ne i} \sum_{j=1}^{V} \mu_{li}  \eta_{lj} \log (\frac{\eta_{lj}}{\eta_{ij}})
\end{equation}
subject to
\begin{equation}
\begin{aligned}
& \sum_{j=1}^{V} \eta_{ij} = 1 \\
\end{aligned}
\end{equation}

The updating direction $\Delta \eta_i$ with a feasible starting point $\eta_i^{0}$ can be calculated using the following equation
\begin{equation}
\begin{bmatrix}
\nabla ^2g_i  & \mathbf{1} \\
\mathbf{1}^T & 0

\end{bmatrix}
\begin{pmatrix}
\Delta \eta_i \\
\alpha_i
\end{pmatrix}
= \begin{pmatrix}
-\nabla g_i \\
0
\end{pmatrix}
\end{equation}
The updating direction is
\begin{equation}
\Delta \eta_{ij} = \frac{- g^\prime_{ij} - \alpha_i }{g^{\prime \prime}_{ij}} 
					  = \frac{ \eta_{ij} (\sum_{d=1}^{D}\sum_{n=1}^{N_d}\phi_{dni}w_{dn}^j  - \sum_{l \ne i} \mu_{li}\eta_{lj}) + \eta_{ij}^2 \big( \sum_{l \ne i} \mu_{il} (\log \eta_{ij} - \log \eta_{lj} + 1) - \alpha_i \big)}
					  { \sum_{d=1}^{D} \sum_{n=1}^{N_d} \phi_{dni} w_{dn}^j   - \sum_{l \ne i} \mu_{li}\eta_{lj} - \eta_{ij} \sum_{l \ne i} \mu_{il}}
\end{equation}
where $g^\prime_{ij}$ and $g^{\prime \prime}_{ij}$ are the first and second partial derivatives of $g_i$ with respect to $\eta_j$, and
\begin{equation}
\alpha_i = \frac{-\sum_{j=1}^V g^\prime_{ij}/g^{\prime \prime}_{ij}}{\sum_{j=1}^V 1 /g^{\prime \prime}_{ij}} = \frac{\sum_{j=1}^{V}\frac{ \eta_{ij} (\sum_{d=1}^{D}\sum_{n=1}^{N_d}\phi_{dni}w_{dn}^j  - \sum_{l \ne i} \mu_{li}\eta_{lj}) + \eta_{ij}^2 \big( \sum_{l \ne i} \mu_{il} (\log \eta_{ij} - \log \eta_{lj} + 1)  \big)}
	{ \sum_{d=1}^{D} \sum_{n=1}^{N_d} \phi_{dni} w_{dn}^j   - \sum_{l \ne i} \mu_{li}\eta_{lj} - \eta_{ij} \sum_{l \ne i} \mu_{il}} }{\sum_{j=1}^{V}\frac{\eta_{ij}^2}{ \sum_{d=1}^{D} \sum_{n=1}^{N_d} \phi_{dni} w_{dn}^j   - \sum_{l \ne i} \mu_{li}\eta_{lj} - \eta_{ij} \sum_{l \ne i} \mu_{il}}}
\end{equation}
The Newton decrement is
\begin{align*}
\lambda(\eta_i)  &= (\Delta \eta_i^T \nabla^2g_i \Delta \eta_i)^{1/2}   \\					
						&= \left( \sum_{j=1}^{V} \frac{ \Big(\sum_{d=1}^{D} \sum_{n=1}^{N_d}\phi_{dni}w_{dn}^j  - \sum_{l \ne i} \mu_{li}\eta_{lj} + \eta_{ij} \big( \sum_{l \ne i} \mu_{il} (\log \eta_{ij} - \log \eta_{lj} + 1) - \alpha_i \big) \Big) ^2}
						{ \sum_{d=1}^{D} \sum_{n=1}^{N_d} \phi_{dni} w_{dn}^j   - \sum_{l \ne i} \mu_{li}\eta_{lj} - \eta_{ij} \sum_{l \ne i} \mu_{il}} \right)^{1/2} 					    
\end{align*}
Due to the non-convexity, $\Delta \eta_i^T \nabla^2g_i \Delta \eta_i$ could be negative at some points. When it happens, the Hessian descent is invoked and finds a new position along $\Delta \eta_i$ with smaller values (see details in Algorithm \ref{kl_algo}). 

To update all the topics, we optimize the topics sequentially and stops until an overall convergence measured by the Frobenius norm of the successive updates, as in Algorithm \ref{kl_algo}. 

\begin{algorithm}[H] \label{kl_algo}
	\KwResult{Update the topic word distribution $\eta$ }
	Initialize the topic word distribution $\eta^0$ with a feasible point for every topic $\eta^0_i, i=1, \dots, K$\;
	Choose the stopping criterion $\epsilon$\;
	\While{$|| \eta^{t+1} - \eta^t ||_F > \epsilon $}{
		\For{topic $i, i=1, \dots, K$}{
			\eIf{$\Delta \eta_i^T \nabla^2f_i \Delta \eta_i \ge 0$}{
			\textbf{Gradient Descent}: \newline
			 update topic word distribution $\eta_i | \eta_j, j \ne i$ using algorithm \ref{pointwise_algo}
		}{\textbf{Hessian Descent}: \newline
		 find step size $h$ satisfying $\eta_i + h\Delta \eta_i > 0$ and $\eta_i - h\Delta \eta_i > 0$\;
		  \eIf{$g_i(\eta_i + h\Delta \eta_i) > g_i(\eta_i - h\Delta \eta_i)$}{$\eta_i \coloneqq \eta_i - h\Delta \eta_i$}{$\eta_i \coloneqq \eta_i + h\Delta \eta_i$}}
}}
	\caption{Variational m-step for the pairwise KL-Divergence}
\end{algorithm}

\iffalse
The KL divergence penalty is in someway connected to the elementwise LASSO penalty. The penalty in equation \ref{equ:kl_per_topic} can be written as follows, if we let  $\mu_{il} = \mu,  \forall i, l$,
\begin{equation}
	\mu \sum_{j=1}^{V} \sum_{l \ne i}(\eta_{ij} - \eta_{lj})(\log \eta_{ij} - \log \eta_{lj})
\end{equation} 
If $\eta_{ij}$ is close to $\eta_{lj}$ for some $l \ne i$, then the $l$th term has no contribution to the minimization. The minimization procedure would make $\eta_{ij}$ as different to $\eta_{lj}, \forall l \ne i$ as possible. 
\fi

\subsection{Dynamic penalty weight implementation and the combination of two penalties} \label{sec:dpw}
Both Algorithm \ref{pointwise_algo} and \ref{kl_algo} apply to the variational m-step. Recall that the ELBO is maximized by an iterative variational EM algorithm. The m-step depends on the e-step output, i.e. $\phi_{dni}w_{dn}^j$ in equation \ref{equ:pw_min} and \ref{equ:kl_per_topic}. In each iteration, E-step would possibly produce  $\phi_{dni}w_{dn}^j$  of different scales, especially at the beginning of the iterations. A penalty weight appropriate for the current iteration may be too big(small) for the next iteration. Therefore we reparameterize the penalty weight $\mu_i$ as $\nu_i * \max_j(\sum_{dn}\phi_{dni}w_{dn}^j )$ in the final variational EM algorithm. The reparameterization makes the penalty similar scale as its ELBO part, and thus effective in every EM iteration.

The combination of two penalties in a single algorithm is straightforward. Algorithm \ref{kl_algo} can be used for the combined penalties. Adding the weighted lasso penalty changes the updating direction $\Delta \eta$ and the Newton decrement $\lambda(\eta)$ in the gradient descent, as it alters the first derivative by subtracting the weights.

\section{Simulation} \label{sec:simulation}
In this section, we use simulated data to demonstrate the effectiveness of the proposed ETM. In section \ref{sec:fw}, we simulate the situation that the corpus contains several frequently appearing words. Comparing to LDA, ETM effectively avoids the frequent word dominance issue in the estimated topics. In section \ref{sec:sim2}, we simulate the case in which field-important words appear infrequently in the underlying corpus. ETM is able to reveal the importance of these words and recover the true distribution using negative weights on these words and zero weights on all the other words, while LDA is not. In section \ref{sec:sim3}, we simulate a 'close' estimated topic situation, by adding several frequently appearing words. LDA topics share these frequently appearing words in their corresponding top-$T$ words. If practitioners interpret topics based on these top-$T$ words, they might mistakenly interpret them as the same topic. ETM is able to separate the topics and at the same time recovers the true topics. 

\subsection{Case 1: Corpus-specific common words} \label{sec:fw}
The setup is as follows. The number of topics $K$ is 2. The prior $\zeta = 0.1$. Topic word distribution $\eta$ is randomly drawn from a Dirichlet prior $\eta \sim Dir(0.1 * \mathbbm{1})$ where $\mathbbm{1}$ is a $300 \times 1$ vector of $1$s. Then we use the word generating process of LDA to generate the words for a corpus containing $500$ documents. During the word generation, we set an upper limit on the maximum number of words in every document to be $100$. Some words don't appear in the corpus due to small topic word probabilities assigned to them. Thus the number of generated words is 202. We further assume that the corpus contains 3 corpus-specific common words 301,  302, 303 and each of them randomly appears in $50\%$ of the documents.  The appearance frequency in a document is 3. These words are then added to the generated corpus. In total, we have $205$ words (202 generated words + 3 manually inserted words) in our final corpus. 

We apply the LDA and ETM ($\nu = 0$)  to the simulated corpus. The weights are the document frequencies of all words scaled to a maximum of 100. The penalty weight is selected to be $\mu = 0.3$.  In practice, the estimated topics are interpreted using their corresponding top-$T$ words. We list the top 10 words of true topics, LDA estimated topics, and ETM topics in Table \ref{tab:sim1}. The corpus-specific common words 301, 302, 303 are assigned high probabilities in LDA and appear in the top 10 words. The probabilities of these 3 words in ETM are about half of those in LDA (a further reduction is achievable with a larger penalty). The high probabilities assigned to these frequently appearing words not only make the topic interpretation difficult, but also distort the document topic frequencies $\theta$, and thus reduce the accuracy of information retrieval.   

\begingroup
\thispagestyle{empty}
\begin{table}[htbp]
	\centering
	\ra{1.1}
	\small
	\begin{tabular*}{\columnwidth} {@{\extracolsep{\fill}} lrrrrrrrrrr}
		\toprule
		Topic 1 & \multicolumn{10}{l}{Top 10 words}  \\
		\midrule
		True & 				47&   83&   86 & 81& 153& 270& 80& 14& 291& 258\\
		LDA & 				47& \textbf{303}& \textbf{302}& \textbf{301}& 83& 86& 81& 270& 153& 80 \\
		ETM &  47&   86&   83& 153& 81& 14& 258& 30& 270& 196 \\
		
		\midrule
		Topic 2 & \multicolumn{10}{l}{Top 10 words}  \\
		\midrule
		True & 170& 256& 206& 0& 219& 286& 243& 114& 132& 82 \\
		LDA & 170& 206& 256& 0& \textbf{301}& \textbf{303}& \textbf{302}& 219& 243& 114\\
		ETM & 170& 206& 256& 219& 243& 114& 0& 286& 82& 132\\
		\bottomrule
	\end{tabular*}%
	
	\caption{\small Top 10 words of the true topics, LDA estimated topics and ETM estimated topics of the simulation study. The corpus-specific frequently appearing words \textbf{301, 302, 303} appear in the LDA estimated topics, but not in the ETM estimated topics. The large probabilities assigned to these frequently appearing words \textbf{301, 302, 303} not only make the topic interpretation difficult but also distort the document topic distribution $\theta$, which reduces the accuracy of information retrieval. The weights are the document frequencies of all words scaled to a maximum of 100. The penalty weight is selected to be 0.3. } \label{tab:sim1}%
\end{table}%
\endgroup

We repeat the above procedure 1000 times and record the number of corpus-specific common words appearing in the top 10 words and the ratio of the average probabilities of these three words in ETM and those in LDA. The summary statistics are given in Table \ref{tab:sim2}. Row 1 is the summary statistics for the number of common words appearing in the top 10 words of LDA topics. The minimum and maximum are 4 and 6, respectively. The mean is 5.989 and the variance is 0.013. It indicates that these 3 corpus-specific common words almost always appear in the top 10 words of LDA estimated topics. Row 2 lists the summary statistics of the number of common words appearing in the top 10 words of document frequency ETM. Its minimum and maximum are 0 and 6, respectively. The mean is 0.585 and the variance is 1.410. It indicates that the majority of the simulation gets no common words in the top 10 words of the ETM estimated topics. The last row shows the summary statistics of the ratio of average probabilities of these three corpus-specific common words in ETM and those in LDA. The minimum and maximum are 0.113 and 0.547. The mean is 0.322 and the variance is 0.004. It indicates that on average, the probabilities assigned to these corpus-specific common words in the ETM are about 1/3 of those in LDA.

\begin{table}[htbp]
	\centering
	\ra{1.1}
	\small
	\begin{tabular*}{\columnwidth} {@{\extracolsep{\fill}} lrrrr}
		\toprule
		  				& min & max & mean & variance  \\
		\midrule
		LDA: number of common words & 		 4& 6 & 5.989 & 0.013  \\
		ETM: number of common words &  0&   6&   0.585& 1.410 \\
		Ratio of common words average probabilities  & 0.113 & 0.547 & 0.322 &0.004 \\

		\bottomrule
	\end{tabular*}%
	
	\caption{\small Summary statistics of 1000 repetition. The first row reports the statistics of the number of common words appearing in the top 10 words of LDA topics. The average is 5.989 and the variance is 0.013, which indicates that these three common words almost always appear in every LDA topic. The second row reports the statistics of the number of common words appearing in the top 10 of ETM topics. The average is 0.585 and the variance is 1.410, indicating the majority don't have the common words in their top 10. The last row lists the statistics of the ratio of the average probabilities of these three common words in ETM and that in LDA. The mean is 0.322 and the variance is 0.004, indicating that on average the probabilities assigned to these three common words in ETM are about 1/3 of those in LDA. } \label{tab:sim2}%
\end{table}%

\subsection{Case 2: Important words appear rarely in the corpus} \label{sec:sim2}
We consider another situation. In practice, certain words are important for that field. But unfortunately, they appear rarely in the current corpus. Practitioners might perceive these words as very important and would like them to be assigned high probability in the topic word distribution. The LDA word generating process is the same as before. Namely, the corpus contains 2 topics. The prior $\zeta = 0.1$. Topic word distribution $\eta$ is randomly drawn from a Dirichlet prior $\eta \sim Dir(0.1 * \mathbbm{1})$ where $\mathbbm{1}$ is a $300 \times 1$ vector of $1$s. We assume that due to some reason, the top 2 words in each topic appear rarely in the current corpus. They only appear in $10\%$ of the total documents, i.e. we randomly select 50 documents containing the top words and delete them from the remaining documents containing them. 

We then apply LDA and ETM ($\nu=0$) to these words. Different from the previous setup in which corpus-specific common words are not known and we use document frequencies as weights, these important words are known and we assign negative weights to them and zero weights to all the other words. In the simulation, words 275 and 22 are important for topic 1, and words 73, 195 are important for topic 2. Their total appearance is limited to 50 documents, i.e. $10\%$ of the corpus size. Due to their rare appearance, LDA is unable to recover their importance and small probabilities are assigned to them. They don't appear in the top 10 words of LDA topics. By assigning weight -100 to these four words and penalty weight $\mu = 0.2$, ETM successfully recovers their position in the top 10 words of the estimated topics. 

\begin{table}[htbp]
	\centering
	\ra{1.1}
	\small
	\begin{tabular*}{\columnwidth} {@{\extracolsep{\fill}} lrrrrrrrrrr}
		\toprule
		Topic 1 & \multicolumn{10}{l}{Top 10 words}  \\
		\midrule
		True & 				\textbf{275}&\textbf{22}&110&251&291&151&171&18&253&187 \\
		LDA & 				110&251&151&291&171&253&18&187&35&287 \\
		ETM & \textbf{275}&\textbf{22}&110&251&151&291&171&253&18&187 \\
		
		\midrule
		Topic 2 & \multicolumn{10}{l}{Top 10 words}  \\
		\midrule
		True & \textbf{ 73}&\textbf{195}&294&207&48&248&19&211&43&175 \\
		LDA & 294&207&48&248&19&43&211&175&269&213\\
		ETM & \textbf{73}&\textbf{195}&294&207&48&248&19&43&211&175\\
		\bottomrule
	\end{tabular*}%
	
	\caption{\small Top 10 words of the true topics, LDA estimated topics and ETM estimated topics of the simulation study. We assume that words 275, 22, 73, 195 are important words of the field, but appears rarely in the current corpus. Because of their rare appearance, LDA is unable to recover them and restore their importance in the topic word distribution. ETM is able to capture the importance of these words, by assigning negative weights to them and zero weights to all the other words. The field important words are usually known in advance by practitioners. Here we assign weights -100 to  words 275, 22, 73, 195, and 0 to all the other words. The penalty ratio is 0.2. } \label{tab:sim3}%
\end{table}%

We repeat the above procedure 1000 times. The summary statistics are reported in Table \ref{tab:sim4}. Row 1 reports the number of important words appearing in the top 10 of LDA estimated topics. The minimum and maximum are 0 and 3, respectively. The mean is 0.299 and the variance is 0.288. It indicates that these important but rarely appearing words barely appear in the top 10 words of LDA topics, i.e. LDA is unable to restore their importance. Row 2 reports the number of important but infrequent words appearing in the top 10 of ETM topics. The minimum and maximum are 2 and 4, respectively. The mean is 3.993 and the variance is 0.009. It indicates that these words are almost always recovered by the ETM. The last row reports the statistics of the average ratio between the probabilities assigned to these important but rarely appearing words in the ETM and those in the LDA. The minimum and maximum are 4.187 and 24.857, respectively. The average is 9.767 and the variance is 6.427. It indicates that on average the probabilities assigned to these important but rarely appearing words are about 10 times in the ETM than those in LDA. 

\begin{table}[htbp]
	\centering
	\ra{1.1}
	\small
	\begin{tabular*}{\columnwidth} {@{\extracolsep{\fill}} lrrrr}
		\toprule
		& min & max & mean & variance  \\
		\midrule
		LDA: number of rare important words & 		 0& 3 & 0.299 & 0.288  \\
		LASSO LDA: number of rare important words &  2&   4&   3.993& 0.009 \\
		Ratio of rare important words average probabilities  & 4.187& 24.853 & 9.767&6.427 \\
		
		\bottomrule
	\end{tabular*}%
	
	\caption{\small Summary statistics of 1000 repetition. The first row reports the statistics of the number of rarely appearing but important words appearing in the top 10 words of LDA topics. The average is 0.299 and the variance is 0.288, which indicates that these important but rarely appearing words seldom appear in the top 10 words of LDA estimated topics. The second row reports the statistics of the number of important but rarely appearing words appearing in the top 10 of ETM topics. The average is 3.993 and the variance is 0.009, indicating the ETM successfully recovers the important words. The last row lists the statistics of the ratio of the average probabilities of these important but rarely appearing words in ETM and that in LDA. The mean is 9.767 and the variance is 6.427, indicating that on average the probabilities assigned to these words in ETM are about 10 times of those in LDA. } \label{tab:sim4}%
\end{table}%

\subsection{Case 3: `Close' topics} \label{sec:sim3}
Practitioners often find some estimated topics are 'close' to each other. By 'close', we mean the estimated topics share several common words and have similar semantic meaning. The exact reason for this phenomenon is unclear. We hypothesize that it is due to the exchangeability assumption of words in LDA. Nevertheless, we simulate the case using frequently appearing words. The basic set up is similar to Case 1. Namely, the corpus contains 2 topics. The prior $\zeta = 0.1$. Topic word distribution $\eta$ is randomly drawn from a Dirichlet prior $\eta \sim Dir(0.1 * \mathbbm{1})$ where $\mathbbm{1}$ is a $300 \times 1$ vector of $1$s. To simulate the estimated 'close' topics in LDA, we further add 6 common words 301, 302, 303, 304, 305, 306 to the corpus and assume that they appear in $80\%$ of the documents. With this setup, we apply LDA and ETM ($\mu=0$) with penalty weight for the pairwise KL divergence $\nu=0.5$ to the simulated corpus. The true and estimated topics are listed in Table \ref{tab:sim5}. Because of the dominance of frequently appearing words, the LDA topics are 'close' to each other, as by our design. The ETM clearly separates them. Although the appearing sequence of ETM is slightly different from the true model, the number of same words appearing in both true and ETM are 8 for both topic 1 and 2, while that for true topics and LDA are 4 and 5 for topic 1 and 2 respectively. The Jensen-Shannon divergence of the true, LDA, and ETM topics are $0.81, 0.63, 0.88$, respectively.  
\begin{table}[htbp]
	\centering
	\ra{1.1}
	\small
	\begin{tabular*}{\columnwidth} {@{\extracolsep{\fill}} lrrrrrrrrrr}
		\toprule
		Topic 1 & \multicolumn{10}{l}{Top 10 words}  \\
		\midrule
		True & 			196& 56& 166& 122& 219& 161& 18& 104& 276& 86	 \\
		LDA & 			196& 56& 166& 122& 301& \textbf{303}& \textbf{306}& \textbf{302}& \textbf{305}& \textbf{304} \\
	    ETM &    134& 56& 196& 122& 161& 166& 86& 219& 104& 301 \\
		
		\midrule
		Topic 2 & \multicolumn{10}{l}{Top 10 words}  \\
		\midrule
		True & 46& 165& 115& 140& 53& 280& 138& 19& 174& 290\\
		LDA & 46& 165& 115& \textbf{304}& 140& \textbf{305}& \textbf{302}& \textbf{306}& 280& \textbf{303} \\
		ETM & 85& 46& 165& 140& 115& 280& 53& 19& 304& 138\\
		\bottomrule
	\end{tabular*}%
	
	\caption{\small Top 10 words of the true topics, LDA estimated topics and ETM estimated topics of the simulation study. Due to the frequently appearing words, LDA topics are 'close' to each other. Although the ETM produce the wrong word sequence, they share 8 words with the true topics for both topic 1 and 2, while LDA only has 4 and 5 shared words for topic 1 and 2. Moreover, the Jensen-Shannon divergence of the true topics, topics estimated by LDA, and topics estimated by ETM are  $0.81, 0.63, 0.88$, respectively.} \label{tab:sim5}%
\end{table}%

We repeat the simulation 1,000 times and report the summary statistics in Table \ref{tab:sim6}. The first column shows the number of shared top 10 words between topics 0 and topic 1. The true topics on average share $0.34$ words with a standard deviation of $0.57$. LDA estimated topics on average share $3.10$ words with a standard deviation of $1.61$. ETM on average share $0.08$ words with a standard deviation of $0.30$. It shows the ETM is capable of separating the 'close' topics and making them share few words in their top words. While the first column focus on the top words, the second column is on the overall topic distribution. It reports the Jensen-Shannon Divergence(JSD) of the topic distributions. The JSD of the true topic is on average $0.80$ with a standard deviation of $0.05$, while LDA is on average $0.64$ with a standard deviation of $0.04$, and ETM is on average $0.79$ with a standard deviation of $0.06$. It shows the ETM separates the 'close' topics. The last two columns report the number of shared top 10 words between the true topics and estimated topics. It doesn't make sense if the ETM separates topics but makes the estimation far away from the true topics. Because of the setup, LDA topic and the true topics on average share $5.54$ and $5.50$ words with standard deviations being $1.24$ and $1.19$ for topics 0 and 1, respectively. ETM and the true topics share on average $8.23$ and $8.20$ words with standard deviation being $1.25$ and $1.28$, respectively. It means that judging from the top-$T$ words, the ETM topics are semantically close to the true model.     

	\begin{table}[htbp]
		\centering
		\ra{1.1}
		\small
		%\caption{Add caption}
		\begin{tabular*}{\columnwidth} {@{\extracolsep{\fill}} lcccc}
			\toprule
			& \multicolumn{1}{p{3cm}}{\# of shared top 10 words between topics 0 and 1} & \multicolumn{1}{l}{JSD} & \multicolumn{1}{p{2.5cm}}{\# of shared top 10 words between true topic 0 and estimated topic 0} & \multicolumn{1}{p{3cm}}{\# of shared top 10 words between true topic 1 and estimated topic 1} \\
			\midrule
			\multirow{2}[0]{*}{True} & 0.34  & 0.80  &       &  \\ [-0.2cm]
			& (0.57)  & (0.05)  &       &  \\
			\multirow{2}[0]{*}{LDA} & 3.10  & 0.64  & 5.54  & 5.50 \\ [-0.2cm]
			& (1.61)  & (0.04)  & (1.24)  & (1.19) \\
			\multirow{2}[0]{*}{KL Div} & 0.08  & 0.79  & 8.23  & 8.20 \\ [-0.2cm]
			& (0.30)  & (0.06)  & (1.25)  & (1.28) \\
			\bottomrule
		\end{tabular*}%
		\small
		\caption{\small Summary statistics of 1000 repetition. The first column shows the mean and standard deviation of the number of shared top 10 words between topics 0 and 1.The  values show that ETM is capable of separating the 'close' topics and their top words share few common words. The second column reports the mean and standard deviation of the corresponding Jensen-Shannon Divergence. On average, the ETM topics are separated and their average JSD is similar to the true topics. The last two columns report the number of shared words between the true topic and the estimated topic. ETM on average shares about $8.2$ words with the true model, meaning the ETM topics are close to the true topics.}\label{tab:sim6}%
	\end{table}%

\section{Real Data Application} \label{sec:nips}
To test the empirical performance of our proposed method, we apply LDA, ETM with lasso penalty only($\nu=0$), and ETM with pairwise KL divergence only ($\mu=0$) to the NIPS dataset, which consists of 11,463 words and 7,241 NIPS conference paper from 1987 to 2017. The data is randomly split into two parts: training (80\%) and testing (20\%). We select the number of topics for LDA using cross-validation with perplexity on the training dataset \citep{LDA}. The selected number is assumed to be the true number of topics in the NIPS dataset. The candidates are $\{5, 10, 15, 20, 25, 30\}$. $K=10$ produces the lowest average validation perplexity. For ETM, we use homogeneous hyperparameters in this experiment, i.e. $\mu_i = \mu, \forall i \in \{1, \dots, K\} $ for the weighted lasso penalty and $\nu_{il} = \nu, \forall i, l \in \{ 1, \dots, K\} \text{ and } l \ne i$ for the pairwise KL divergence penalty, as we don't have any prior information on the topics. 

We do cross-validation on the training data to select the penalty weight $\mu$. When selecting $\mu$, perplexity is no longer an appropriate measure. Perplexity is the negative likelihood per word. A higher probability of frequently appearing words will produce a lower perplexity. As a result, $\mu = 0$ will be selected. Another commonly used metric to select hyperparameters is the topic coherence score. Researchers have proposed several calculation methods of the topic coherence scores \citep{newman2010automatic, mimno2011optimizing, aletras2013evaluating, roder2015exploring}. \cite{ roder2015exploring} show that among all these proposed topic coherence scores, $C_V$ achieves the highest correlation with all available human topic ranking data (also see \cite{syed2017full}). Roughly speaking, $C_V$ takes into consideration both generalization and localization of the topics. Generalization means $C_V$ measures the performance on the unseeable test dataset. Localization means $C_V$ uses a rolling window to measure the word's co-occurrence.

In this paragraph, we provide the details of  $C_V$ calculation. The top $N$ words of each topic are selected as the representation of the topic, denoted as $W = \{w_1, \dots, w_N\}$. Each word $w_i$ is represented by an $N$-dimensional vector $v(w_i) = \{NPMI(w_i, w_j) \}_{j=1, \dots, N}$, where $j$th-entry is the Normalized Pointwise Mutual Information(NPMI) between word $w_i$ and $w_j$, i.e. $NPMI(w_i, w_j) = \frac{\log P(w_i, w_j)  - \log (P(w_i)P(w_j))}{-\log P(w_i, w_j)}$. $W$ is represented by the sum of all word vectors, $v(W) = \sum_{j=1}^N v(w_j)$. The calculation of NPMI between word $w_i$ and $w_j$ involves the marginal and joint probabilities $p(w_i), p(w_j), p(w_i, w_j)$. A sliding window of size 110, which is the default value in the python package 'gensim' and robust for many applications,  is used to create pseudo-document and estimate the probabilities. The purpose of the sliding window is to take the distance between two words into consideration. For each word $w_i$, a pair is formed $(v(w_i), v(W))$. A cosine similarity measure $\phi_i(v(w_i), v(W)) = \frac{v(w_i)^T v(W)}{\norm{v(w_i)} \norm{v(W)} }$ is then calculated for each pair. The final $C_V$ score for the topic is the average of all $\phi_i$s.

We use $C_V$ to select the penalty weight from  $\{0, 0.5, 1, 1.5, 2, 2.5, 3\}$.  $\mu=0.5$ produces the highest average coherence score 0.56 for ETM, while $\mu=0$, i.e. LDA, gives the coherence score 0.51. We then refit both LDA and ETM ($\mu=0.5, \nu=0$) to the whole training dataset and use the test dataset to calculate the coherence score $C_V$ as a final evaluation of the performance on unseeable data. The results are shown in Table \ref{tab:dflda}. Overall the $C_V$ score of LDA topics is 0.51 and that of ETM is 0.62, a $22\%$ improvement. We highlight two common words 'data' and 'using' in the top 20 words of both topics. They appear more frequently in LDA topics than in ETM topics. Both words appear in 5 out of 10 LDA topics and 1 out of 10 ETM topics. We also observe some large improvements for topic \textit{Reinforcement Learning, Neural Network}, \textit{Computer Vision}, and several other topics. Take \textit{Reinforcement Learning} as an example. Comparing the top 20 words, we observe that words \textit{time, value, function, model, based, problem} appear in LDA topic, but not the ETM topic. Surely these words are associated with reinforcement learning, but they are also associated with topics \textit{Neural Network, Bayesian, Optimization, etc}. We refer these words as corpus-specific common words, i.e. for the current corpus, they contain little information to distinguish one topic from another. Their positions in the ETM topic are filled by words \textit{game, trajectory, robot, control}. These words are related to the applications of reinforcement learning and represent the topics better than the previous corpus-specific common words. We see the $C_V$ score increased from 0.56 to 0.77, a $38\%$ increase. We do observe that an LDA topic related to NLP is missing in ETM. One possible reason is that along the way of variational EM algorithm, the algorithm converges to different points for this topic. One way to avoid this is to initialize the ETM with a rough estimate from LDA.

\begin{table}[htbp]
	\centering
	\ra{1.3}
	\footnotesize
	\begin{tabular*}{\columnwidth} {@{\extracolsep{\fill}}p{2cm} p{13cm}r}
		%\begin{tabular}{llllllllllllllllllllrr}
		\toprule
		Topics & Top 20 words & $C_V$ \\
		\midrule 
		LDA   &       & 0.51 \\
		\midrule
		Machine Learning &  matrix \textbf{data} kernel problem algorithm sparse linear method rank methods \textbf{using} dimensional analysis vector space function norm error matrices set & 0.42 \\
		\textit{Reinforcement Learning} &  \textit{state learning policy action \underline{time} \underline{value} reward \underline{function} \underline{model} optimal actions states agent control reinforcement algorithm \textbf{using} \underline{based} decision \underline{problem}} & \textit{0.56} \\
		\textit{\pbox[l][10pt]{2cm}{Neural\\ Network}} &  \textit{model time neurons figure spike neuron neural response stimulus activity visual input information cells signal fig cell noise brain synaptic} & \textit{0.65} \\
		\textit{Computer Vision} &  \textit{image images learning model training deep \textbf{using} layer neural object network networks features recognition use models dataset feature results different} & \textit{0.57} \\
		NLP   &  model word models words features \textbf{data} set figure \textbf{using} human topic speech object language objects used recognition context based feature & 0.50 \\
		\pbox[l][10pt]{2cm}{Neural\\ Network} &  network networks neural input learning output training units hidden error weights time function weight layer figure number set used memory & 0.52 \\
		Bayesian &  model distribution \textbf{data} models log gaussian likelihood bayesian inference parameters posterior prior \textbf{using} process distributions latent variables mean time probability & 0.49 \\
		\pbox[l][10pt]{2cm}{Graph\\ models}  &  graph algorithm tree set clustering node nodes number cluster structure problem \textbf{data} time variables graphs edge clusters random algorithms edges & 0.52 \\
		Optimization &  algorithm bound theorem log function learning let algorithms bounds problem convex loss optimization case set convergence functions optimal gradient probability & 0.43 \\
		Classification &  learning \textbf{data} training classification set class test error examples function classifier \textbf{using} label feature features loss problem kernel performance svm & 0.46 \\
		\midrule
		ETM &       & 0.62 \\
		\midrule
		Machine Learning &  matrix rank sparse pca tensor lasso subspace spectral manifold norm matrices recovery sparsity eigenvalues kernel principal eigenvectors singular entries embedding & 0.55 \\
		\textit{Reinforcement Learning}  &  \textit{policy action reward agent state actions reinforcement policies \underline{game} agents states \underline{trajectory} \underline{robot} planning \underline{control} \underline{trajectories} rewards \underline{games} exploration transition} & \textit{0.77} \\
		\textit{\pbox[l][10pt]{2cm}{Neural\\ Network}} &  \textit{neurons network neuron spike input neural synaptic time firing activity dynamics output networks fig circuit spikes cell signal analog patterns} & \textit{0.71} \\
		\textit{Computer Vision} &  \textit{image images object objects segmentation scene pixel face detection video pixels vision patches visual shape recognition motion color pose patch} & \textit{0.78} \\
		\pbox[l][10pt]{2cm}{Neural\\ Network} &  model visual stimulus brain response spatial human stimuli responses subjects motion frequency cells temporal cortex signals signal activity filter motor & 0.73 \\
		\pbox[l][10pt]{2cm}{Neural\\ Network} &  layer network deep networks units hidden word layers training convolutional trained neural speech recognition architecture language recurrent net input output & 0.73 \\
		Bayesian &  inference latent posterior tree variational bayesian node models topic nodes variables model likelihood markov distribution graphical gibbs prior dirichlet sampling & 0.54 \\
		\pbox[l][10pt]{2cm}{Graph\\ models}  &  convex graph algorithm optimization clustering gradient convergence problem theorem solution algorithms dual descent submodular stochastic iteration graphs objective max problems & 0.50 \\
		Optimization &  bound theorem regret loss bounds algorithm risk lemma log let proof online ranking bounded bandit query setting hypothesis complexity learner & 0.52 \\
		Classification &  learning \textbf{data} model set \textbf{using} function algorithm number time figure given results training used based problem error models use distribution & 0.37 \\
		\bottomrule
\end{tabular*}%
\footnotesize
\caption{\small Top 20 words of the topics estimated by LDA and ETM.} \label{tab:dflda}%
\end{table}%

\begin{table}[htbp]
	\centering
	\ra{1.3}
	\footnotesize
	\begin{tabular*}{\columnwidth} {@{\extracolsep{\fill}}p{2cm} p{13cm}r}
	%\begin{tabular}{llllllllllllllllllllrr}
	\toprule
	Topics & Top 20 words & $C_V$ \\
	\midrule 
	LDA & & 0.52 \\
	\midrule
	\textit{Machine Learning} &	\textit{matrix  \underline{data}   kernel  sparse  linear  \underline{points}  \underline{problem}  rank   algorithm  space  \underline{using}  dimensional  \underline{method}  analysis  matrices  vector  clustering  error  \underline{set}    \underline{methods}}        & \textit{0.42} \\
	\textit{Reinforcement Learning}	&\textit{state  \underline{learning}  policy  action  \underline{time}   reward  \underline{value}  \underline{\textbf{function}}  algorithm  optimal  agent  actions  states  reinforcement  \underline{problem}  control  \underline{\textbf{model}}  decision  \underline{using}  \underline{based}}        & \textit{0.56} \\
	\pbox[l][10pt]{2cm}{Neural\\ Network}	&\textbf{model}  time   neurons  figure  spike  neuron  neural  information  response  activity  stimulus  visual  cells  cell   input  fig    signal  brain  different  synaptic        & 0.65 \\
	\textit{Computer Vision} &	\textit{image  images  \underline{\textbf{model}}  object  \underline{training}  deep   \underline{using}  \underline{learning}  features  recognition  \underline{models}  layer  feature  figure  objects  \underline{use}    visual  \underline{different}  \underline{results}  vision}        & \textit{0.55} \\
	Theory &	theorem  bound  algorithm  let    log    learning  \textbf{function}  probability  bounds  loss   case   distribution  error  set    proof  lemma  functions  following  sample  given        & 0.42 \\

	Bayesian &	\textbf{model}  data   distribution  models  gaussian  log    likelihood  parameters  using  posterior  bayesian  prior  inference  process  latent  \textbf{function}  mean   time   distributions  sampling        & 0.47 \\
	Graphical Models &	graph  tree   \textbf{model}  node   nodes  set    algorithm  structure  number  variables  models  inference  graphs  clustering  cluster  edge   edges  figure  time   topic        & 0.48 \\
	\pbox[l][10pt]{2cm}{Neural\\ Network} &	network  neural  networks  input  learning  output  training  units  layer  hidden  time   weights  error  figure  \textbf{function}  weight  used   set    using  state   &      0.52 \\
	Optimization &	algorithm  optimization  gradient  problem  \textbf{function}  convex  algorithms  method  methods  convergence  solution  learning  set    time   objective  problems  linear  stochastic  step   descent        & 0.54 \\
	Classification & 	learning  data   training  classification  set    features  feature  using  class  \textbf{model}  test   task   classifier  label  used   based  performance  examples  number  labels        & 0.55 \\
		\midrule
	ETM& &0.57    \\
		\midrule
	\textit{Machine Learning} &	\textit{clustering  kernel  matrix  norm   rank   kernels  \underline{spectral}  \underline{pca}    matrices  tensor  subspace  \underline{lasso}  \underline{manifold}  \underline{eigenvalues}  \underline{embedding}  \underline{principal}  \underline{singular}  completion  recovery  \underline{eigenvalue}}        & \textit{0.54} \\
	\textit{Reinforcement Learning}	&	\textit{action  policy  reward  agent  actions  state  \underline{game}   reinforcement  \underline{regret}  arm    \underline{planning}  policies  \underline{exploration}  \underline{robot}  \underline{games}  agents  \underline{player}  states  bandit  rewards}         & \textit{0.76} \\
	\pbox[l][10pt]{2cm}{Neural\\ Network}	&	neural  input  time   figure  \textbf{model}  neurons  visual  neuron  fig    spike  response  information  spatial  signal  activity  pattern  cell   different  temporal  cells        & 0.64 \\
	\textit{Computer Vision} &	\textit{\underline{faces}  image  images  layer  object  deep   \underline{segmentation}  layers  \underline{convolutional}  objects  \underline{pixel}  \underline{scene}  \underline{pixels}  \underline{video}  architecture  recognition  vision  networks  \underline{face}   pose}         & \textit{0.73} \\
	Theory &	bound  algorithm  theorem  let    \textbf{function}  learning  log    set    case   probability  bounds  functions  loss   error  following  proof  problem  given  optimal  random        & 0.40 \\
	\pbox[l][10pt]{2cm}{Neural\\ Network} &	gates  network  units  networks  sonn   recurrent  net    hidden  architecture  layer  analog  feedforward  backpropagation  chip   nets   connectionist  gate   modules  module  feed        &  0.60* \\
	Bayesian &	data   \textbf{model}  gaussian  distribution  prior  models  log    mean   parameters  likelihood  noise  estimation  estimate  density  using  variance  bayesian  mixture  samples  process        & 0.47 \\
	Graphical Models &	graph  \textbf{model}  models  tree   nodes  node   inference  structure  variables  number  markov  set    graphs  edge   time   topic  edges  probability  cluster  graphical       & 0.47 \\
	Optimization &	algorithm  optimization  problem  algorithms  gradient  method  methods  solution  \textbf{function}  convergence  objective  problems  step   linear  iteration  stochastic  max    update  descent  learning        & 0.54 \\
	Classification &	learning  data   classification  training  set    feature  test   features  task   class  classifier  label  using  examples  performance  used   labels  tasks  based  word         & 0.57 \\
		\bottomrule
	\end{tabular*}%
	\footnotesize \\
	\emph{The $0.60^*$ is computed by removing the word \textit{sonn}. As \textit{sonn} doesn't appear in the testing dataset, we get NaN for the $C_V$ score.} \\
	\caption{\small Top 20 words of the topics estimated by LDA and ETM.} \label{tab:rlda}%
\end{table}%

For the ETM with only pairwise KL divergence penalty ($\mu=0$), due to the non-convexity in the M-step optimization, we initialize the topic using an estimation of LDA. Coincidentally, $\nu=0.5$ also produces the largest coherence score 0.53, and $\nu=0$ gets 0.50. Same as before, we refit both models to the whole training dataset and compute their $C_V$ score using the testing dataset as a final evaluation. The results are shown in Table \ref{tab:rlda}. Overall the topic estimated by LDA has a coherence score of 0.52, while that of ETM is 0.57. As a way to measure the 'distance' of the estimated topics, we calculate the Jensen-Shannon Divergence(JSD) of both topics. The JSD of LDA topics is 0.93, while that of ETM topics is 1.90. From a distance point of view, the ETM topics are more separated from each other. For the NIPS dataset, we don't observe similar topics. Although the third and eighth topics are both interpreted as neural network, they emphasize different aspects of the topic. The third topic is from a biological point of view. It contains words \textit{spike, neuron, stimulus, brain, synatic}. The eighth topic is of computer science point of view. It contains words \textit{network, learning, training, output, layer, hidden}. When separating the estimated topics, ETM suppresses the appearance of less topic relevant words and improves the topic coherence. For example, LDA \textit{Machine Learning} topic contains words \textit{points, problem, using, set, methods}. ETM suppresses the appearance of these words. Instead, it promotes words \textit{spectral, pca, lasso, manifold, eigenvalues, embedding, principal, singular}, which are better representatives of the topic. As a result, the $C_V$ score improves from 0.42 to 0.54, a $29\%$ improvement. Similarly, LDA \textit{Reinforcement Learning} topic contains words \textit{learning, time, value, function, problem, model, using, based}, which are kind of common and can have high probabilities in other topics, e.g. \textit{Neural Network, Computer Vision, Theory, Optimization, etc}. ETM replace those words by more specific and related words \textit{game, regret, planning, exploration, robot}. The $C_V$ score increases from 0.56 to 0.76, a $36\%$ improvement. The same goes for the topic \textit{Computer Vision}. Words \textit{model, training, using, learning, use, different, results} are suppressed in ETM. Words \textit{faces, segmentation, convolutional, pixel, video} which are unique to the topic are promoted in the ETM topic. The $C_V$ score increases from 0.55 to 0.73, a $33\%$ improvement. Although some corpus-specific common words are suppressed under both penalties, the underlying reasons are different. Under the weighted lasso penalty, common words are penalized because they appear in too many documents. Under the pairwise KL divergence penalty, some common words are suppressed because they appear in other topics. To make topics distant from each other, the pairwise KL divergence penalty suppresses their appearance in less relevant topics.

\section{Conclusion} \label{sec:conclusion}
Motivated by the frequently appearing words dominance in the discovered topics, we propose an Exclusive Topic Model (ETM), which contains a weighted lasso penalty term and a pairwise KL divergence term. The penalties destroy the close form solution for the topic distribution as in LDA. Instead, we estimate the topics using constrained Newton's method for the case of having the weighted lasso penalty only and a combination of gradient descent and Hessian descent for having the pairwise KL divergence only. The combination of gradient descent and Hessian descent algorithm is ready to be applied for the ETM with both penalties, with a little twist to the gradient of the objective function. Although the intention is to solve the frequent words intrusion issue for the weighted lasso penalty, it is not limited to the sole purpose. Practitioners can utilize the weights to incorporate their prior knowledge to the topics. We demonstrate the effectiveness of the proposed model using three simulation studies, where in each case ETM is superior to LDA and recovers the true topics. We also apply the proposed method to the publicly available NIPS dataset. Compare with LDA, our proposed method assign lower weights to the commonly appearing words, making the topics easier to interpret. The topic coherence score $C_V$ also shows that topics are more semantically consistent than those estimated from LDA. 

The ETM with only a weighted LASSO penalty is related to \cite{bhattacharya2015dirichlet}. They claim that the Dirichlet-Laplace priors possess optimal posterior concentration and lead to efficient posterior computation. The LASSO penalties can be viewed as the Laplace prior in the posterior. The weights control the mixture between Dirichlet prior for the topic drawing and Laplace prior. In our current setup, the topic distributions are treated as estimated parameters. With the Dirichlet-Laplace prior, we can adopt the full Bayesian approach that the topics are generated from the Dirichlet-Laplace prior. We would reach a very similar posterior with our current setup, except for an extra variational distribution for the topics. Instead of estimating the topic parameters, we would estimate the variational parameters for the topics. The benefit of the full Bayesian approach is that we can make use of the optimal posterior concentration property \citep{bhattacharya2015dirichlet} and theoretical properties of variational inference \citep{yang2020alpha, zhang2020convergence, pati2018statistical, wang2019frequentist} to show some properties of the proposed method. On the other hand, it is more challenging to derive the theoretical properties related to the pairwise KL divergence penalty, as there is no ready prior distribution corresponding to it. The non-convexity means that we are only able to obtain a local minimum for the topic distributions.

\bibliography{ResearchProposalLDAReference.bib}
\end{document}